%% file: main.tex
\documentclass[runningheads]{llncs}
\usepackage{subfiles}
\usepackage{graphicx}
\usepackage{amsmath,amssymb} % define this before the line numbering.
\usepackage{color}
\usepackage{booktabs}
\usepackage{bbm}
\usepackage{ftnxtra}
\usepackage{fnpos} % \makeFNbelow by default
\usepackage[font=small,skip=0pt]{caption}
\usepackage[width=122mm,left=12mm,paperwidth=146mm,height=193mm,top=12mm,paperheight=217mm]{geometry}
\usepackage{wrapfig}
\begin{document}
\subfile{eccv2018submission.tex}

\subfile{supp.tex}
\end{document}

%% file: eccv2018submission.tex
% \linenumbers
\pagestyle{headings}
%\mainmatter
%\def\ECCV18SubNumber{1818}  % Insert your submission number here

\title{Audio-Visual Event Localization in Unconstrained Videos} % Replace with your title

% \titlerunning{Audio-Visual Event}

% \authorrunning{Audio-Visual Event}

% \author{Yapeng Tian}
% \institute{University of Rochester}
\titlerunning{Audio-Visual Event Localization in Unconstrained Videos}
\authorrunning{Y. Tian, J. Shi, B. Li, Z. Duan, and C. Xu}

\author{Yapeng Tian$^1$, Jing Shi$^1$, Bochen Li$^{2}$,
       Zhiyao Duan$^2$, and Chenliang Xu$^1$
       }
\institute{$^1$Department of Computer Science, University of Rochester\\
$^2$Department of Electrical and Computer Engineering, University of Rochester
}

\maketitle

\begin{abstract}
%Where is the mosquito? Listen first, then look for it. This depicts people's real-life experience in locating mosquito at home, where vision and audition work cohesively leads to a success. However, does it apply to machines in understanding unconstrained videos? In this paper, we aim to provide a systematic study for answering the above question.Observing the lack of literature, 
In this paper, we introduce a novel problem of audio-visual event localization in unconstrained videos. We define \textit{an audio-visual event} as an event that is both visible and audible in a video segment. We collect an \textit{Audio-Visual Event} (AVE) dataset
to systemically investigate three temporal localization tasks: supervised and weakly-supervised audio-visual event localization, and cross-modality localization. We develop an audio-guided visual attention mechanism to explore audio-visual correlations, propose a dual multimodal residual network (DMRN) to fuse information over the two modalities, and introduce an audio-visual distance learning network to handle the cross-modality localization. Our experiments support the following findings: joint modeling of auditory and visual modalities outperforms independent modeling, the learned attention can capture semantics of sounding objects, temporal alignment is important for audio-visual fusion, the proposed DMRN is effective in fusing audio-visual features, and strong correlations between the two modalities enable cross-modality localization. 
%video dataset of 4143 clips spanning over 28 audio-visual events 
%\ZD{Why is spatial localization relevant here? Isn't the paper about temporal localization? - the paper is about temporal localization but we introduce a attention which can capture the semantics of sounding objects} \ZD{Yes, I would state this as additional contributions. Basically, separate it into multiple sentences. (- we can only write 150 words in abstract, so I put all of these together)}
%\ZD{What is ``cross-modality localization''? Temporal localization? - given audio, locate its corresponding visual segment and vice versa. We show task in Fig. 1 and problems.} \ZD{Ok. This sounds to be closer to the core task than the spatial attention contribution.}
%\textbf{at least 70 and at most 150 words and now 214 words} 
%\cxu{Get rid of the mosquito example in title and abstract, and perhaps the dataset. Update the findings with your new experiments.}
%e.g., given audio, locate its corresponding visual segment and vice versa. 
%\dots
\end{abstract}

\section{Introduction}

% Why it is important to do audio-visual, some works are in constrained domain. 
Studies in neurobiology suggest that the perceptual benefits of integrating visual and auditory information are extensive~\cite{bulkin2006seeing}. For computational models, they reflect in lip reading~\cite{DBLP:journals/corr/AssaelSWF16,ChSeViCVPR2017}, where correlations between speech and lip movements provide a strong cue for linguistic understanding; in music performance~\cite{LiXuDuSMC2017}, where vibrato articulations and hand motions enable the association between sound tracks and the performers; and in sound synthesis~\cite{owens2016visually}, where physical interactions with different types of material give rise to plausible sound patterns. Albeit these advances, these models are limited in their constrained domains.  

% More recent ones are in unconstrained domain, but we have a different focus. 
Indeed, our community has begun to explore marrying computer vision with audition \textit{in-the-wild} for learning a \textit{good} representation \cite{aytar2016soundnet,owens2016ambient,Arandjelovic2017ICCV}. For example, a sound network is learned in~\cite{aytar2016soundnet} by a visual teacher network with a large amount of unlabeled videos, which shows better performance than learning in a single modality. However, they have all assumed that the audio and visual contents in a video are matched (which is often not the case as we will show) and they are yet to explore whether the joint audio-visual representations can facilitate understanding unconstrained videos.

\begin{figure}[t!]
\setlength\belowcaptionskip{-20pt}
\begin{center}
  \includegraphics[width=\linewidth]{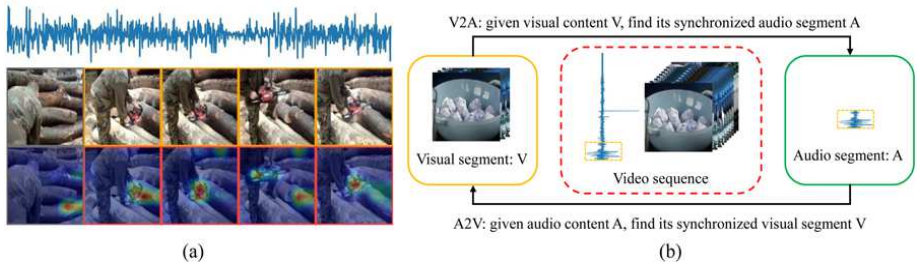}
\end{center}
\caption
{(a) illustrates audio-visual event localization. The first two rows show a 5s video sequence with both audio and visual tracks for an audio-visual event \textit{chainsaw} (event is temporally labeled in yellow boxes). The third row shows our localization results (in red boxes) and the generated audio-guided visual attention maps. (The first frame does not contain the chainsaw event, hence the attention focuses on background regions.) (b) illustrates cross-modality localization for V2A and A2V 
%\ZD{Swapping the yellow and green boxes may make the figure easier to understand, i.e., audio on the left and video on the- it is cross-modal localization, so V is near to the audio sequence}  
}
\label{fig:tasks}
\end{figure}

% What do we care about? The audio-visual scene understanding. 
In this paper, we study a family of audio-visual event temporal localization tasks (see Fig. \ref{fig:tasks}) as a proxy to the broader audio-visual scene understanding problem for unconstrained videos. We pose and seek to answer the following questions: 
(Q1) Does inference jointly over auditory and visual modalities outperform inference over them independently? 
(Q2) How does the result vary under noisy training conditions? 
(Q3) How does knowing one modality help model the other modality?
%(Q3) How does one modality help attend the other modality? 
%\ZD{This question is a little vague.} 
(Q4) How do we best fuse information over both modalities? 
(Q5) Can we locate the content in one modality given its observation in the other modality?
%(Q5) Do the correlations existing between the two modalities support cross-modality localization?
%\ZD{What correlations? this seems to be the first time mentioning it. Also, what is cross-modal localization? First time mentioning.} 
Notice that the individual questions might be studied in the literature, but we are not aware of any work that conducts a systematic study to answer these collective questions as a whole. 

% How do we answer these questions? Define first tasks. 
In particular, we define an \textit{audio-visual event} as an event that is both visible and audible in a video segment, and we establish three tasks to explore aforementioned research questions: 1) supervised audio-visual event localization, 2) weakly-supervised audio-visual event localization, and 3) event-agnostic cross-modality localization. The first two tasks aim to predict which temporal segment of an input video has an audio-visual event and what category the event belongs to. The weakly-supervised setting assumes that we have no access to the temporal event boundary but an event tag at video-level for training. Q1-Q4 will be explored within these two tasks. In the third task, we aim to locate the corresponding visual sound source temporally within a video from a given sound segment and vice versa, which will answer Q5.

% Method 
We propose both baselines and novel algorithms to solve the above three tasks. For the first two tasks, we start with a baseline model treating them as a sequence labeling problem. We utilize CNN \cite{lecun1998gradient} to encode audio and visual inputs, adapt LSTM \cite{hochreiter1997long} to capture temporal dependencies, and apply Fully Connected (FC) network to make the final predictions. 
Upon this baseline model, we introduce an audio-guided visual attention mechanism to verify whether audio can help attend visual features; it also implies spatial locations for sounding objects as a side output. Furthermore, we investigate several audio-visual feature fusion methods and propose a novel dual multimodal residual fusion network that achieves the best fusion results. For weakly-supervised learning, we formulate it as a Multiple Instance Learning (MIL) \cite{maron1998framework} task, and modify our network structure via adding a MIL pooling layer to handle the problem. To address the harder cross-modality localization task, we propose an audio-visual distance learning network that measures the relativeness of any given pair of audio and visual content. It projects audio and visual features into subspaces with the same dimension. Contrastive loss \cite{hadsell2006dimensionality} is introduced to learn the network. 

% Dataset & Findings
Observing that there is no publicly available dataset directly suitable for our tasks, we collect a large video dataset that consists of 4143 10-second videos with both audio and video tracks for 28 audio-visual events and annotate their temporal boundaries. Videos in our dataset are originated from YouTube, thus they are unconstrained. 
%We design proper metrics to evaluate the three localization tasks. 
Our extensive experiments support the following findings: modeling jointly over auditory and visual modalities outperforms modeling independently over them, audio-visual event localization in a noisy condition can still achieve promising results, the audio-guided visual attention can well capture semantic regions covering sounding objects and can even distinguish audio-visual unrelated videos, temporal alignment is important for audio-visual fusion, the proposed dual multimodal residual network is effective in addressing the fusion task, and strong correlations between the two modalities enable cross-modality localization. These findings have paved a way for our community to solve harder, high-level understanding problems in the future, such as video captioning~\cite{XuMeYaCVPR2016} and movieQA~\cite{TaZhStCVPR2016}, where the auditory modality plays an important role in understanding video but lacks effective modeling. 

% Contributions 
Our work makes the following contributions: (1) a family of three audio-visual event localization tasks; (2) an audio-guided visual attention model to adaptively explore the audio-visual correlations; (3) a novel dual multimodal residual network to fuse audio-visual features; (4) an effective audio-visual distance learning network to address cross-modality localization; (5) a large audio-visual event dataset containing more than 4K unconstrained and annotated videos, which to the best of our knowledge, is the largest dataset for sound event detection. We will release our dataset along with implementations of various methods. %upon acceptance. 

\section{Related Work}
\label{sec:related}

In this section, we first describe how our work differs from three closely-related topics: sound event detection, temporal action localization and multimodal machine learning, then discuss relations to various recent works in modeling vision-and-sound.

Sound event detection considered in the audio signal processing community aims to detect and temporally locate sound events in an acoustic scene. Approaches based on Hidden Markov Models (HMM), Gaussian Mixture Models (GMM), feed-forward Deep Neural Networks (DNN), and Bidirectional Long Short-Term Memory (BLSTM)~\cite{schuster1997bidirectional} are developed in~\cite{heittola2013context,mesaros2016tut,cakir2015polyphonic,parascandolo2016recurrent}. These methods focus on audio signals, and visual signals have not been explored. Corresponding datasets, e.g., TUT Acoustic Scenes~\cite{7760424}, for sound event detection only contain sound tracks, and are not suitable for audio-visual scene understanding.

Temporal action localization aims to detect and locate actions in videos. Most works cast it as a classification problem and utilize a temporal sliding window approach, where each window is considered as an action candidate subject to classification~\cite{oneata2013action}. Escorcia~\emph{et al.} \cite{escorcia2016daps} present a deep action proposal network that is effective in generating temporal action proposals for long videos and can speed up temporal action localization. Recently, Shou~\emph{et al.} \cite{Shou_2016_CVPR} propose an end-to-end Segment-based 3D CNN method (S-CNN), and Lea~\emph{et al.} \cite{lea2016temporal} develop an Encoder-Decoder Temporal Convolutional Network (ED-TCN) to hierarchically model actions. Different from these works, an audio-visual event in our consideration may contain multiple actions or motionless sounding objects, and we model over both audio and visual domains. Nevertheless, we extend the ED-TCN method to address our supervised audio-visual event localization task and compare it in Sec. \ref{evc}. 

Multimodal machine learning aims to learn joint representations over multiple input modalities, e.g., speech and video, image and text. Feature fusion is one of the most important part for multimodal learning \cite{baltruvsaitis2018multimodal}, and many different fusion models have been developed, such as statistical models \cite{fisher2001learning}, Multiple Kernel Learning (MKL) \cite{gonen2011multiple,poria2015deep}, Graphical models \cite{gurban2008dynamic,ngiam2011multimodal}. Although some mutimodal deep networks have been studied in \cite{ngiam2011multimodal,srivastava2012multimodal,srivastava2012learning,mroueh2015deep,hu2016temporal,Yang_2017_CVPR,kiela2018efficient}, which mainly focus on joint audio-visual representation learning based on Autoencoder or deep Boltzmann machines \cite{srivastava2012multimodal}, we are interested in investigating the best models to fuse learned audio and visual features for localization purpose.

Recently, some inspiring works are developed for modeling vision-and-sound \cite{Arandjelovic2017ICCV,aytar2016soundnet,owens2016ambient,owens2016visually,harwath2016unsupervised}. Aytar~\emph{et al.} \cite{aytar2016soundnet} use a visual teacher network to learn powerful sound representations from unlabeled videos. Owens~\emph{et al.} \cite{owens2016ambient} leverage ambient sounds as supervision to learn visual representations. Arandjelovic and Zisserman \cite{Arandjelovic2017ICCV} learn both visual and audio representations in an unsupervised manner through an audio-visual correspondence task, and in \cite{arandjelovic2017objects}, they further locate sound source spatially in an image based on an extended correspondence network. Aside from works in representation learning, audio-visual cross-modal synthesis is studied in \cite{owens2016ambient,Yipin18,Lele17},   
and associations between natural image scenes and accompanying free-form spoken audio captions are explored in \cite{harwath2016unsupervised}. 
Unlike the previous works, in this paper, we systematically investigate audio-visual event localization tasks.

%------------------------------------------
\section{Dataset and Problems}
%\section{AVE: The \emph{Audio-Visual Event} Dataset}
\label{sec:dataset}
%------------------------------------------
%\subsection{Gathering and Preparing Dataset}
\subsection{AVE: The \emph{Audio-Visual Event} Dataset}
\begin{figure}[t!]
\setlength\belowcaptionskip{-20pt}
\begin{center}
  \includegraphics[width=\linewidth]{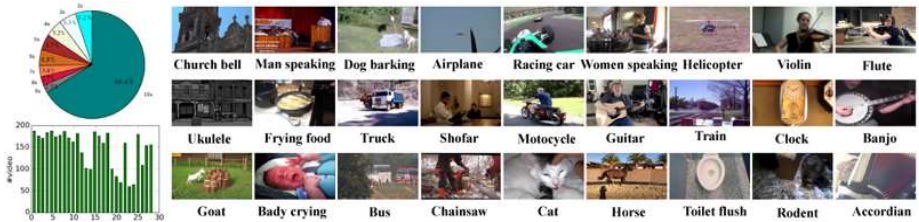}
\end{center}
   \caption{The AVE dataset. Some examples in the dataset are shown. The distribution of videos in different categories and the distribution of event lengths are illustrated}
\label{fig:dataset}
\end{figure}

To the best of our knowledge, there is no publicly available dataset directly suitable for our purpose. Therefore, we introduce the \emph{Audio-Visual Event} (AVE) dataset\footnote{The supplementary material contains the detail of gathering the dataset.}, a subset of AudioSet \cite{gemmeke2017audio}, that contains 4143 videos covering 28 event categories and videos in AVE are temporally labeled with audio-visual event boundaries. 
%-----------------------------------------
%Our AVE dataset contains in total 4143 videos. 
Each video contains at least one 2s long \textit{audio-visual event}. The dataset covers a wide range of audio-visual events (\emph{e.g.}, man speaking, woman speaking, dog barking, playing guitar, and frying food \emph{etc.}) from different domains, e.g., human activities, animal activities, music performances, and vehicle sounds. We provide examples from different categories and show the statistics in Fig.~\ref{fig:dataset}. Each event category contains a minimum of 60 videos and a maximum of 188 videos, and 66.4$\%$ videos in the AVE contain audio-visual events that span over the full 10 seconds. Next, we introduce three different tasks based on the AVE to explore the interactions between auditory and visual modalities.
%YouTube is a good source for finding unconstrained videos. 
%AudioSet \cite{gemmeke2017audio} released by Google is a large-scale video dataset of many manually-annotated audio events---it consists of 632 audio events and a collection of 2M human-labeled 10-second clips drawn from YouTube. It contains a variety of human and animal sounds, musical instruments and genres, and common everyday environmental sounds. Although videos in AudioSet contain both audio and visual tracks, a lot of them are not suitable for audio-visual event localization. For example, visual and audio content can be completely unrelated (\emph{e.g.}, train horn but no train appears, wind sound but no corresponding visual signals, the absence of audible sound, \emph{etc}). 

%We select around $10,000$ videos covering 34 categories from the AudioSet. We hire trained annotators to choose a subset of more appropriate videos and mark the start and end time of each audio-visual event at a time resolution of 1 second. We set a standard that all annotators followed to choose and annotate videos: a desired appropriate video should include an audio-visual event segment with at least two seconds long, in which the sound source is visible and the sound is audible. The finally selected subset contains 4143 videos covering 28 event categories.

%------------------------------------------
%\subsection{Dataset Statistics}

%-----------------------------------------
%\section{Problems}
%\label{sec:problems}
\subsection{Fully and Weakly-Supervised Event Localization}
\label{sec:data:avel}

The goal of event localization is to predict the event label for each video segment, which contains both audio and visual tracks, for an input video sequence. Concretely, for a video sequence, we split it into $T$ non-overlapping segments $\{V_{t}, A_{t}\}_{t = 1}^{T}$, where each segment is 1s long (since our event boundary is labeled at second-level), and $V_{t}$ and $A_{t}$ denote the visual content and its corresponding audio counterpart in a video segment, respectively. Let $\textbf{\textit{y}}_t = \{y_{t}^k|y_{t}^k \in \{0, 1\}, k = 1, ..., C, \sum_{k=1}^{C}y_{t}^k = 1\}$ be the event label for that video segment. Here, $C$ is the total number of AVE events plus one background label.

For the supervised event localization task, the event label $\textbf{\textit{y}}_t$ of each visual segment $V_t$ or audio segment $A_t$ is known during training. We are interested in event localization in audio space alone, visual space alone and the joint audio-visual space. This task explores whether or not audio and visual information can help each other improve event localization. Different than the supervised setting, in the weakly-supervised manner we have only access to a video-level event tag, and we still aim to predict segment-level labels during testing. The weakly-supervised task allows us to alleviate the reliance on well-annotated data for  modelings of audio, visual and audio-visual.

%\subsection{Weakly-Supervised Event Localization}

%Let $S$ be a video, and within the video there are $N$ audio-visual segment pairs $\{V_{i}, A_{i}\}_{i = 1}^{N}$. In the supervised event localization problem, each training pair $\{V_{i}, A_{i}\}$ has its label $\textbf{\textit{y}}_i$. 
%However, in the weakly-supervised manner, label $\textbf{\textit{y}} = \{y^{k}|y^{k} \in \{0, 1\}, k = 1, ..., C, \sum_{k=1}^{C}y^{k} = 1\}$ is only available at the video level in the training phase, and we do not know segment-level label $\textbf{\textit{y}}_i$. Specifically, if the label of $S = \{S_i\}_{i=1}^{N}$ is $\textbf{\textit{y}}$, we only know at least one $S_i$ belongs to the event category, but we cannot confirm which segment pairs are background or contain the event. However, in the testing phase, we would still care about the event prediction performance at the segment-level.

\subsection{Cross-Modality Localization}
\label{sec:data:cross}

In the cross-modality localization task, given a segment of one modality (auditory/visual), we would like to find the position of its synchronized content in the other modality (visual/auditory). 
Concretely, for visual localization from audio (A2V), given a $l$-second audio segment $\hat{A}$ from $\{A_t\}_{t=1}^{T}$, where $l < T$, we want to find its synchronized $l$-second visual segment within $\{V_t\}_{t=1}^{T}$. Similarly, for audio localization from visual content (V2A), given a $l$-second video segment $\hat{V}$ from $\{V_t\}_{t=1}^{T}$, we would like to find its $l$-second audio segment within $\{A_t\}_{t=1}^{T}$. 
This task is conducted in the event-agnostic setting such that the models developed for this task are expected to work for general videos where the event labels are not available. For evaluation, we only use short-event videos, in where the lengths of audio-visual event are all shorter than 10s.

%Note that the absolute temporal boundaries of the event in the given modality is not provided to the system and the system needs to rely on the audio-visual correlation to localize the event in the target modality. 
%It allows us to explore the deeper question: whether a model trained with audio-visual information can learn to locate information in one modality given the observation of the other modality within same video. 

% \cxu{The current prob. def. is ill posed.} To maintain a concise writing, we overwrite the definitions in the previous section. Let $N$ training samples be $\{V_i, A_i, y_i\}_{i=1}^{N}$, where $V_i$ and $A_i$ are a pair of visual and audio inputs, $y_i \in \{0, 1\}$ is their label. Here, $y_i = 1$ means that $V_i$ and $A_i$ are synchronized and they describe the same event content, i.e., a true-matched pair, otherwise, $y_i = 0$. 

%The cross-modality localization problem is event category-independent. In this task, synchronized audio and video pairs are labeled as positive. However, the positive visual or auditory pairs can include various different events unlabeled in the unconstrained condition. Therefore, this problem is very challenging.  

%------------------------------------------
\section{Methods for Audio-Visual Event Localization}
\label{sec:avel}
%=======================================
\begin{figure}[t!]
\begin{center}
  \includegraphics[width=0.9\linewidth]{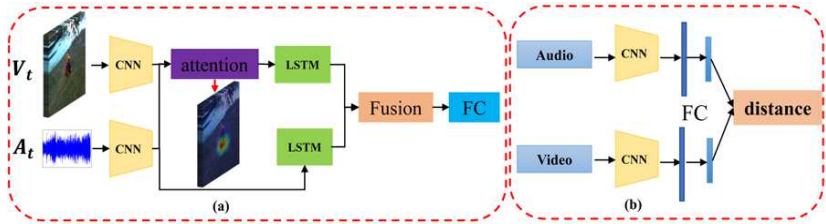}
\end{center}
   \caption{(a) Audio-visual event localization framework with audio-guided visual attention and multimodal fusion. One timestep is illustrated, and note that the fusion network and FC are shared for all timesteps. (b) Audio-visual distance learning network}
\label{fig:framework}
\end{figure}
% \cxu{If space is a problem, we may want to cut some equations as they are not that special. Only show when necessary. Many equations are just a duplicate of each other in a different context, we would like to find a way to avoid that.}

First, we present the overall framework that treats the audio-visual event localization (defined in Sec.~\ref{sec:data:avel}) as a sequence labeling problem in Sec.~\ref{sec:avel:base}. Upon this framework, we propose our audio-guided visual attention in Sec.~\ref{sec:avel:attention} and a novel dual multimodal residual fusion network in Sec.~\ref{sec:avel:fusion}. Finally, we extend this framework to work in weakly-supervised setting in Sec.~\ref{sec:avel:weak}. 

\subsection{Audio-Visual Event Localization Network}
\label{sec:avel:base}
%\vspace{2mm}
%\noindent \textbf{Audio-Visual Network} 

%\cxu{Describe a base CNN+LSTM network here independent of attention and various fusion method. You will introduce them as model variants in 5.2 and 5.3. --(various compared fusion methods are covered in other papers and I introduce them in later experiments section)} 
Our network mainly consists of five modules: feature extraction, audio-guided visual attention, temporal modeling, multimodal fusion and temporal labeling (see Fig.~\ref{fig:framework}(a)).
The feature extraction module utilizes pre-trained CNNs to extract visual features $v_{t} = [v_{t}^1, ..., v_{t}^k]\in\mathbb{R}^{d_v\times k}$ and audio features $a_{t}\in\mathbb{R}^{d_a}$ from each $V_{t}$ and $A_{t}$, respectively. Here, $d_v$ denotes the number of CNN visual feature maps, $k$ is the vectorized spatial dimension of each feature map, and $d_a$ denotes the dimension of audio features. We use an audio-guided visual attention model to generate a context vector $v^{att}_t\in\mathbb{R}^{d_v}$ (see details in Sec.~\ref{sec:avel:attention}). Two separate LSTMs take $v^{att}_t$ and $a_{t}$ as inputs to model temporal dependencies in the two modalities respectively. For an input feature vector $F_t$ at time step $t$, the LSTM updates a hidden state vector $h_t$ and a memory cell state vector $c_t$:  
\begin{equation}\label{lstm}
	h_t, c_t = \textrm{LSTM}(F_t, h_{t-1}, c_{t-1})
    \enspace,
\end{equation}
where $F_t$ refers to $v^{att}_t$ or $a_t$ in our model. For evaluating the performance of the proposed attention mechanism, we compare to models that do not use attention; we directly feed global average pooling visual features and audio features into LSTMs as baselines. To better incorporate the two modalities, we introduce a multimodal fusion network (see details in Sec.~\ref{sec:avel:fusion}). The audio-visual representation $h_t^{*}$ is learned by a multimodal fusion network with audio and visual hidden state output vectors $h_t^v$ and $h_t^a$ as inputs. This joint audio-visual representation is used to output event category for each video segment. For this, we use a shared FC layer with the Softmax activation function to predict probability distribution over $C$ event categories for the input segment and the whole network can be trained with a multi-class cross-entropy loss.

\begin{wrapfigure}{r}{0.42\textwidth}
\vspace{-20mm}
\setlength\belowcaptionskip{-25pt}
\begin{center}
\includegraphics[width=0.4\textwidth]{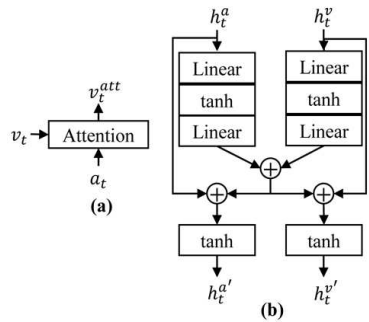}
\end{center}
   \caption{(a) Audio-guided visual attention mechanism. (b) Dual multimodal residual network for audio-visual feature fusion}
\label{fig:fusion}
\end{wrapfigure}

%=========================================
%------------------------------------------
\subsection{Audio-Guided Visual Attention}
\label{sec:avel:attention}

Psychophysical and physiological evidence shows that sound is not only informative about its source but also its location \cite{gaver1993world}. Based on this, Hershey and Movellan \cite{hershey2000audio} introduce an exploratory work on localizing sound sources utilizing audio-visual synchrony. It shows that the strong correlations between the two modalities can be used to find image regions that are highly correlated to the audio signal. Recently, \cite{owens2016ambient,arandjelovic2017objects} show that sound indicates object properties even in unconstrained images or videos. These works inspire us to use audio signal as a means of guidance for visual modeling. 

Given that attention mechanism has shown superior performance in many applications such as neural machine translation \cite{NMT15} and image captioning \cite{xu2015show,lu2017knowing}, we use it to implement our audio-guided visual attention (see Fig.~\ref{fig:framework}(a) and Fig.~\ref{fig:fusion}(a)).
%visual question answering \cite{yang2016stacked,zhao2017video,jang2017tgif}, and image classification \cite{wang2017residual}. 
%For exploiting the expert knowledge that sound indicates sounding objects and better understanding audio-visual correlation in videos, based on recent works in \cite{xu2015show,lu2017knowing,jang2017tgif}, we introduce a mechanism illustrated in Fig. \ref{fig:fusion}(a), which will help understand how auditory modality attends visual modality. 
The attention network will adaptively learn which visual regions in each segment of a video to look for the corresponding sounding object or activity. 

Concretely, we define the attention function $f_{att}$ and it can be adaptively learned from the visual feature map $v_{t}$ and audio feature vector $a_{t}$. At each time step $t$, the visual context vector $v_{t}^{att}$ is computed by: 
\begin{equation}\label{att}
	v_{t}^{att} = f_{att}(a_t, v_t)=\sum_{i = 1}^{k} w_{t}^{i}v_{t}^{i}  
    \enspace,
\end{equation}
where $w_t$ is an attention weight vector corresponding to the probability distribution over $k$ visual regions that are attended by its audio counterpart. The attention weights can be computed based on MLP with a Softmax activation function:
\begin{equation}\label{att_weights}
	w_{t} = Softmax(x_{t}) 
    \enspace,
\end{equation}
\begin{equation}\label{att}
	x_{t} = W_{f}\sigma(W_{v}U_v(v_t) + (W_{a}U_a(a_t))\mathbbm{1}^{T}) 
    \enspace,
\end{equation}
where $U_v$ and $U_a$, implemented by a dense layer with nonlinearity, are two transformation functions that project audio and visual features to the same dimension $d$, $W_v\in\mathbb{R}^{k\times d}$, $W_a\in\mathbb{R}^{k\times d}$, $W_f\in\mathbb{R}^{1\times k}$ are parameters, the entries in $\mathbbm{1}\in\mathbb{R}^{k}$ are all 1, $\sigma(\cdot)$ is the hyperbolic tangent function, and $w_t\in\mathbb{R}^{k}$ is the computed attention map. The attention map visualization results show that the audio-guided attention mechanism can adaptively capture the location information of sound source (see Fig. \ref{fig:att_easy}), and it can also improve temporal localization accuracy (see Tab. \ref{att}).  

%=========================================
%------------------------------------------
\subsection{Audio-Visual Feature Fusion}
\label{sec:avel:fusion}

Our fusion method is designed based on the philosophy in~\cite{srivastava2012multimodal}, which processes multiple features separately and then learns a joint representation using a middle layer.  
To combine features coming from visual and audio modalities, inspired by the Mutimodal Residual Network (MRN) in \cite{kim2016multimodal} (which works for text-and-image), we introduce a Dual Multimodal Residual Network (DMRN). The MRN adopts a textual residual branch and feeds transformed visual features into different textual residual blocks, where only textual features are updated. In contrary, the proposed DMRN shown in Fig.~\ref{fig:fusion}(b) updates both audio and visual features simultaneously. 

Given audio and visual features $h_t^a$ and $h_t^v$ from LSTMs, the DMRN will compute the updated audio and visual features:
\begin{equation}\label{update_a}
	h_t^{a'} = \sigma(h_t^a + f(h_t^a, h_t^v)) 
    \enspace,
\end{equation}
\begin{equation}\label{update_v}
	h_t^{v'} = \sigma(h_t^v + f(h_t^a, h_t^v)) 
    \enspace,
\end{equation}
% \begin{equation}\label{update_h}
% 	h_t^{*} = \frac{1}{2}(h_t^{v'} + h_t^{a'})  
%     \enspace,
% \end{equation}
where $f(\cdot)$ is an additive fusion function, and the average of $h_t^{a'}$ and $h_t^{v'}$ is used as the joint representation $h_t^{*}$ for labeling the video segment. 
Here, the update strategy in DMRN can both preserve useful information in the original modality and add complimentary information from the other modality.
Simply, we can stack multiple residual blocks to learn a deep fusion network with updated $h_t^{a'}$ and $h_t^{v'}$ as inputs of new residual blocks. However, we empirically find that it does not improve performance by stacking many blocks for both MRN and DMRN. We argue that the network becomes harder to train with increasing parameters and one block is enough to handle this simple fusion task well. 

We would like to underline the importance of fusing audio-visual features after LSTMs for our task. We empirically find that late fusion (fusion after temporal modeling) is much better than early fusion (fusion before temporal modeling). We suspect that the auditory and visual modalities are not temporally aligned. Temporal modeling by LSTMs can implicitly learn certain alignments which can help make better audio-visual fusion. The empirical evidences will be shown in Tab. \ref{fusion}.

%=========================================
%------------------------------------------
\subsection{Weakly-Supervised Event Localization}
\label{sec:avel:weak}

To address the weakly-supervised event localization, we formulate it as a MIL problem and extend our framework to handle noisy training condition.
Since only video-level labels are available, we infer label of each audio-visual segment pair in the training phase, and aggregate these individual predictions into a video-level prediction by MIL pooling as in \cite{wu2015deep}:
\begin{equation}\label{bag_rep}
\hat{m} = g(m_{1}, m_{2}, ..., m_{T}) = \displaystyle \frac{1}{T}{\sum_{t = 1}^T m_{t}}
\enspace,
\end{equation}
where $m_1, ..., m_T$ are predictions from the last FC layer of our audio-visual event localization network, and $g(\cdot)$ averages over all predictions. The probability distribution of event category for the video sequence can be computed using $\hat{m}$ over the Softmax. 
During testing, we can predict the event category for each segment according to computed $m_t$.

\section{Method for Cross-Modality Localization}
 \label{sec:cross}
% \begin{figure}[t!]
% \begin{center}
% %\fbox{\rule{0pt}{2in} \rule{0.9\linewidth}{0pt}}
%   \includegraphics[width=0.8\linewidth]{figs/cmln}
% \end{center}
%    \caption{Audio-visual distance learning network. \cxu{merge with fig 2}}
% \label{fig:fusion}
% \end{figure}\cite{hershey2017cnn}
%==========================================

% \cxu{I would consider this an ``new'' contribution for method. Put more emphasizes on this rather than the previous baseline methods in Section 5.}

% \cxu{Watch your commas and periods after equation.}

To address the cross-modality localization problem (defined in Sec.~\ref{sec:data:cross}), we propose an audio-visual distance learning network (AVDLN) as illustrated in Fig. \ref{fig:framework}(b); we notice similar networks are studied in concurrent works~\cite{arandjelovic2017objects,suris2018cross}. Our network can measure the distance $D_{\theta}(V_i, A_i)$ for a given pair of $V_i$ and $A_i$. At test time, for visual localization from audio (A2V), we use a sliding window method and optimize the following objective:  
\newcommand{\argmin}{\arg\!\min}
\begin{equation}\label{avdln_audio_test}
t^* = \argmin_{t} \displaystyle \sum_{s=1}^{l}D_{\theta}(V_{s+t-1}, \hat{A}_s)
\enspace,
\end{equation}
where $t^*\in \{1,..., T-l+1\}$ denotes the start time when visual and audio content synchronize, $T$ is the total length of a testing video sequence, and $l$ is the length of the audio query $\hat{A}$. This objective function computes an optimal matching by minimizing the cumulative distance between the  audio segments and the visual segments.
Therefore, $\{V_i\}_{i=t^*}^{t^*+l-1}$ is the matched visual content. Similarly, we can define audio localization from visual content (V2A); we omit it here for a concise writing. Next, we describe the network used to implement the matching function.

Let $\{V_i, A_i\}_{i=1}^{N}$ be $N$ training samples and $\{y_i\}_{i=1}^{N}$ be their labels, where $V_i$ and $A_i$ are a pair of 1s visual and audio segments, $y_i \in \{0, 1\}$. Here, $y_i = 1$ means that $V_i$ and $A_i$ are synchronized. The AVDLN will learn to measure distances between these pairs. The network encodes them using pre-trained CNNs, and then performs dimensionality reduction for encoded audio and visual representations using two different two-layer FC networks. The outputs of final FC layers are $\{R_i^v, R_i^a\}_{i=1}^{N}$.  
The distance between $V_i$ and $A_i$ is measured by the Euclidean distance between $R_i^v$ and $R_i^a$:
\begin{equation}\label{distance}
D_{\theta}(V_i, A_i) = ||R_i^v - R_i^a||_2
\enspace.
\end{equation}
To optimize the parameters $\theta$ of the distance metric $D_{\theta}$, we introduce the contrastive loss proposed by  Hadsell \emph{et al.} \cite{hadsell2006dimensionality}. The contrastive loss function is: 
\begin{equation}\label{closs}~
L_{C} = y_{i}D_{\theta}^2(V_i, A_i) + (1 - y_i)(\max(0, th - D_{\theta}(V_i, A_i)))^2,
\end{equation}
where $th > 0$ is a margin. If a dissimilar pair's distance is less than $th$, the loss will make the distance $D_{\theta}$ bigger; if their distance is bigger than the margin, it will not contribute to the loss. 
%As validated in \cite{hadsell2006dimensionality}, the contrastive term involving dissimilar pairs, is crucial. If just minimizing the first term in the loss using all similar pairs, it will lead to a collapsed solution (the $D_{\theta}$ will be zero, if setting all network's weights as zeros). The network is optimized by Adam. 

%------------------------------------------
\section{Experiments}
\label{sec:exp}

First, we introduce the used visual and audio representations in Sec.~\ref{representations}. Then, we describe the compared baseline models and evaluation metrics in Sec~\ref{baselines}. Finally, we show and analyze experimental results\footnote{
The supplementary material contains additional results on audio-visual event localization with C3D features, visual-guided audio attention and co-attention, and implementation details of our models.} of different models in Sec.~\ref{evc}.    

\subsection{Visual and Audio Representations}
\label{representations}

%\cxu{Move this to experiment section. This is the detail of experiments that is independent of your methods.}

It has been suggested that CNN features learned from a large-scale dataset (\emph{e.g.} ImageNet \cite{ImageNet}, AudioSet \cite{gemmeke2017audio}) are highly generic and powerful for other vision or audition tasks. So, we adopt pre-trained CNN models to extract features for visual segments and their corresponding audio segments. 
%\cxu{a video voxel is a pixel in video; use video segments; same for one place in your method section.}

\vspace{2mm}
\noindent \textbf{Visual Representation.}  
% Given a 10-seconds video $V$, we split it into 10 non-overlapping segments, and then construct 10 video cubes $\{v_{i}\}_{i = 1}^{N}$ by uniformly sampling $T=16$ video frames from each segment. Therefore, a 1-second video cube $v_{i}$ includes $T$ video frames. 
%Given a 10-second video stream $V$, we split it into $N=10$ non-overlapping segments as video cubes $\{v_{i}\}_{i = 1}^{N}$. Each cube is one-second long containing $T=16$ uniformly sampled frames. 
%To extract visual representation of a 1s video segment, we first extract its spatial visual feature using a CNN for each frame in the segment, and then mean pool the video features across the entire video cube similar to video representation by \cite{venugopalan2014translating} for video captioning. Finally, we can obtain a visual representation $v_t$ for a 1s video segment. 
For each 1s visual segment, we extract $pool5$ feature maps from sampled $16$ RGB video frames by VGG-19 network \cite{SimonyanZ14a}, which is pre-trained on ImageNet \cite{russakovsky2015imagenet}, and then utilize global average pooling \cite{lin2013network} over the 16 frames to generate one $512\times 7\times 7$-D feature map. We also explore the temporal visual features extracted by C3D \cite{tran2015learning}, which is capable of learning spatio-temporal visual features. But we do not observe significant improvements when combining C3D features. 
%\vspace{2mm}

% \noindent  \textbf{Spatial-Temporal Visual Representation.}
% Although 2-D CNNs like VGGNet pre-trained on ImageNet is effective in extracting high-level visual representations for static images, they can not capture dynamic features modeling motion information in videos. To also make use of the motion cues, we utilize deep 3-D convolutional neural network (C3D) \cite{tran2015learning}, which is capable of learning spatio-temporal visual features. In our experiments, for a video cube $v_i$, the feature $F_i^{v_{c3d}}$ extracted by global average pooling the $pool5$ layer of C3D network pre-trained on Sport1M \cite{karpathy2014large} is used. 

\vspace{2mm}
\noindent \textbf{Audio Representation.} We extract a 128-D audio representation for each 1s audio segment via a VGG-like network \cite{hershey2017cnn} pre-trained on AudioSet \cite{gemmeke2017audio}.
%Following work in audio classification \cite{hershey2017cnn}, we decompose an audio segment $a_{i}$ containing $960 ms$ audio frames using a short-time Fourier transform applying $25 ms$ windows every $10 ms$. The resulting spectrogram is integrated into 64 mel-spaced frequency bins, and the magnitude of each bin is logtransformed after adding a small offset to avoid numerical issues. This gives a log-mel spectrogram patch of 96 $\times$ 64 bins for the $960 ms$ audio segment. We extract semantically meaningful, high-level 128-D audio embedding $F_{a}^{i}$ for $a_{i}$ using a pre-trained VGG-like model \cite{hershey2017cnn} from the 96 $\times$ 64 log-mel spectrogram patch.
%%%%tables\hat{a}_s, 

\begin{figure}[t!]
\setlength\belowcaptionskip{-10pt}
\begin{center}
  \includegraphics[width=\linewidth]{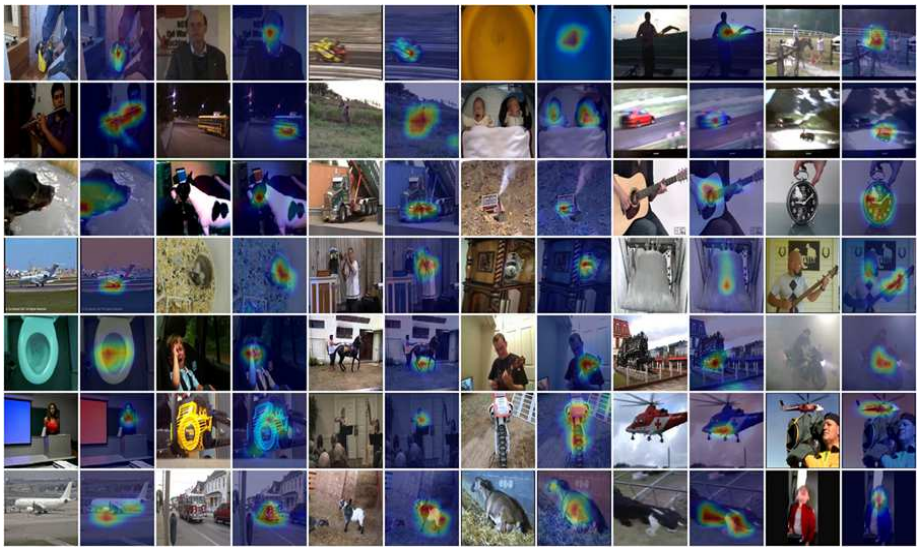}
\end{center}
   \caption{ Qualitative visualization of audio-guided visual attention on AVE test dataset. The semantic regions containing many different sound sources, such as barking dog, crying boy/babies, speaking woman, horning bus, guitar \emph{etc}, can be adaptively captured by our attention model
 }
\label{fig:att_easy}
\end{figure}
\begin{figure}[t!]
\setlength\belowcaptionskip{-10pt}
\begin{center}
  \includegraphics[width=\linewidth]{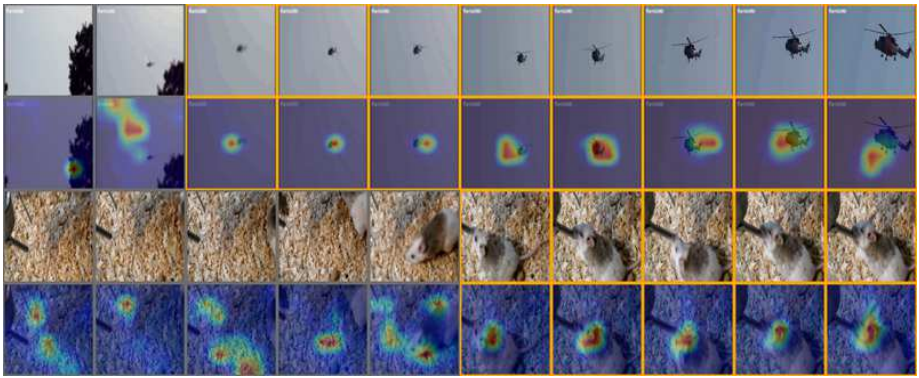}
\end{center}
   \caption{ Visualization of visual attention maps on two \textbf{challenging} examples. The first and third rows are 10 video frames uniformly extracted from two 10s videos, and the second and fourth rows are generated attention maps. The \textbf{yellow} box (groundtruth label) denotes that the frame contain audio-visual event in which sounding object is visible and sound is audible. If there is no audio-visual event in a frame, random background regions that do not focus on objects will be attended (see 2rd frame of first example and 5th frame of the second example); otherwise, the attention will focus on sounding sources   
   }
\label{fig:att_hard}
\end{figure}
%%%%
%------------------------------------------
\subsection{Baselines and Evaluation Metrics}
\label{baselines}

To validate the effectiveness of the joint audio-visual modeling, we use single-modality models as baselines, which only use audio-alone or visual-alone features and share the same structure with our audio-visual models. To evaluate the audio-guided visual attention, we compare our V-att and A+V-att models with V and A+V models in fully and weakly supervised settings. Here, V-att models adopt audio-guided visual attention to pool visual feature maps, and the other V models use global average pooling to compute visual feature vectors. We visualize generated attention maps for subjective evaluation. To further demonstrate the effectiveness of the proposed networks, we also compare them with a state-of-the-art temporal labeling network: ED-TCN \cite{lea2016temporal}.

We compare our fusion method: DMRN with several network-based  multimodal fusion methods: Additive, Maxpooling (MP), Gated, Multimodal Bilinear (MB), and Gated Multimodal Bilinear (GMB) in \cite{kiela2018efficient}, Gated Multimodal Unit (GMU) in \cite{arevalo2017gated}, Concatenation (Concat), and MRN \cite{kim2016multimodal}. Three different fusion strategies: early, late and decision fusions are explored. Here, early fusion methods directly fuse audio features from pre-trained CNNs and attended visual features; late fusion methods fuse audio and visual features from outputs of two LSTMs; and decision fusion methods fuse the two modalities before Softmax layer. 
In addition, to further enhance the performance of DMRN, we also introduce a variant model of DMRN called dual multimodal residual fusion ensemble (DMRFE) method, which feed audio and visual features into two separated blocks and then use average ensemble to combine the two predicted probabilities.

For supervised and weakly-supervised event localization, we use overall accuracy as an evaluation metric. 
For cross-modality localization, e.g., V2A and A2V, if a matched audio/visual segment is exactly the same as its groundtruth, we regard it as a good matching; otherwise, it will be a bad matching. We compute the percentage of good matchings over all testing samples as prediction accuracy to evaluate the performance of cross-modality localization. To validate the effectiveness of the proposed model, we also compare it with deep canonical correlation analysis (DCCA) method \cite{andrew2013deep}. 

%Let $\{V_i, A_i\}_{i=1}^{N}$ be $N$ one-second audio and video segment pairs, $\{y_i\}_{i=1}^{N}$ be its labels, where $y_i \in \{0, 1\}$. Using our cross-modality localization method, we can obtain a predicted label $\{\hat{y}_i\}_{i=1}^{N}$ by labeling the matched segment as 1 and other segments as 0. We compute the labeling precision to evaluate our method. Obviously, precision and recall values will be equal in our method, because the real positive and predicted positive instance numbers are same.    

%------------------------------------------
% \subsection{Implementation Details (\textbf{can we put it in supp.})}
% AVE videos are divided into training (3339), validation (402), and testing (402) sets. For audio-visual event localization tasks, we randomly sample videos from each category to build the train/val/test datasets. For test the cross-modality localization performance, we only sample testing videos from short-event videos (events in these videos are all smaller than 10s). We implement our models using Pytorch \cite{pytorch} and Keras \cite{chollet2015keras} with Tensorflow \cite{tensorflow2015-whitepaper} as backend. Networks are optimized by Adam \cite{kingma2014adam}. 
%LSTM hidden state size is 128. 
%The contrastive loss margin is empirically set to 2.0. 

%------------------------------------------

\setlength{\tabcolsep}{2pt}
\begin{table}[t!]
\setlength\belowcaptionskip{-5pt}
\begin{center}
\caption{Event localization prediction accuracy ($\%$) on AVE dataset. A, V, V-att, A+V, A+V-att denote that these models use audio, visual, attended visual, audio-visual and attended audio-visual features, respectively. W-models are trained in a weakly-supervised manner. Note that audio-visual models all fuse features by concatenating the outputs of LSTMs}
\label{att}
\begin{tabular}{l | c | c c | c c| c| c c |c c}
\toprule
 Models& A  &V    &V-att  &A+V &A+V-att  & W-A &W-V&W-V-att &W-A+V &W-A+V-att\\
\midrule
Accuracy &59.5&55.3&58.6&71.4&\textbf{72.7} &53.4 &52.9 &55.6  &63.7 &66.7\\
\bottomrule
\end{tabular}
\end{center}
\vspace{-5mm}
\end{table}

% %\setlength{\tabcolsep}{2pt}
% \begin{table}[t!]
% \setlength\belowcaptionskip{-5pt}
% \begin{center}
% \begin{tabular}{l | c | c c | c c| c| c c |c c}
% \toprule
%  Models& A  &V    &V-att  &A+V &A+V-att  & W-A &W-V&W-V-att &W-A+V &W-A+V-att\\
% \midrule
% Accuracy &59.5&55.3&58.6&71.4&\textbf{72.7} &53.4 &52.9 &55.6  &63.7 &66.7\\
% \bottomrule
% \end{tabular}
% \end{center}
% \caption{Event localization prediction accuracy ($\%$) on AVE dataset. A, V, V-att, A+V, A+V-att denote that these models use audio, visual, attended visual, audio-visual and attended audio-visual features, respectively. W-models are trained in weakly-supervised manner. Note that audio-visual models fuse audio and visual features by concatenating the outputs of the audio and visual LSTMs.}
% \label{att}
% \end{table}
   
%------------------------------------------
\subsection{Experimental Comparisons}
\label{evc}

Table \ref{att} compares different variations of our proposed models on supervised and weakly-supervised audio-visual event localization tasks. Table \ref{fusion} shows event localization performance of different fusion methods. Figures \ref{fig:att_easy} and \ref{fig:att_hard} illustrate generated audio-guided visual attention maps. 

To benchmark our models with state-of-the-art temporal action labeling methods, we extend the ED-TCN~\cite{lea2016temporal} to address the supervised audio-visual event localization, and train it on AVE. The ED-TCN achieves 46.9\% overall accuracy. For comparison, our V model with the same features achieves 55.3\%.

\vspace{1mm}
\noindent 
\textbf{Audio and Visual.} 
From Tab. \ref{att}, we observe that A outperforms V and W-A is also better than W-V. It demonstrates that audio features are more powerful to address audio-visual event localization task on the AVE dataset. However, when we look at each individual event, using audio is not always better than using visual. We observe that V is better than A for some events (\emph{e.g.}  car, motocycle, train, bus). Actually, most of these events are outdoor. Audios in these videos can be very noisy: several different sounds may be mixed together (\emph{e.g.} people cheers with a racing car), and may have very low intensity (\emph{e.g.} horse sound from far distance). For these conditions, visual information will give us more discriminative and accurate information to understand events in videos. A is much better than V for some events (\emph{e.g.} dog, man and woman speaking, baby crying). Sounds will provide clear cues for us to recognize these events. For example, if we hear barking sound, we know that there may be a dog. We also observe that A+V is better than both A and V, and W-A+V is better than W-A and W-V, which validates that combining audio and visual modalities significantly improve the event localization performance. 

From the above results and analysis, we can conclude that auditory and visual modalities will provide complementary information for us to understand events in videos. The results also demonstrate that our AVE dataset is suitable for studying audio-visual scene understanding tasks.

\vspace{2mm}
\noindent \textbf{Audio-Guided Visual Attention.}
%Our V-att based models utilize audio information to attend visual spatial features adaptively, and the audio-guided visual attention serves as a global pooling method to compute visual feature vectors. 
The quantitative results (see Tab. \ref{att}) show that V-att is much better than V (a 3.3$\%$ absolute improvement) and A+V-att outperforms A+V by 1.3$\%$, which demonstrates the effectiveness of proposed audio-guided visual attention mechanism. We show qualitative results of our attention method in Fig. \ref{fig:att_easy}. We observe that a range of semantic regions in many different categories and examples can be attended by audio, which validates that our attention network can learn which visual regions to look at for sounding objects (even for some challenging cases: two babies crying, playing flute surrounding by crowd, rat with weak sound). An interesting observation is that the audio-guided visual attention tends to focus on sounding regions, such as man's mouth, head of crying boy \emph{etc}, rather than whole objects in some examples. 
Figure \ref{fig:att_hard} illustrates two challenging cases. For the first example, the sounding helicopter is quite small in the first several frames but our attention model can still capture its locations. For the second example, the first five frames do not contain an audio-visual event which means that either the sound source is not visible or sound is not audible; in this case, attentions are spread on different background regions. When the rat appears in the 5th frame but is not making any sound, the attention does not focus on the rat. When the rat sound becomes audible, the attention focuses on the sounding rat. This observation validates that the audio-guided attention mechanism is helpful to distinguish audio-visual unrelated videos, and is not just to capture a saliency map with objects.
%The quantitative and qualitative results demonstrate the effectiveness of the proposed audio-guided attention mechanism.

%------------------------------------------
\setlength{\tabcolsep}{1.4pt}
\begin{table}[t]
\setlength\belowcaptionskip{5pt}
\begin{center}
\caption{Event localization prediction accuracy ($\%$) of different feature fusion methods on AVE dataset. These methods all use same audio and visual features as inputs. Top-2 results in each line are highlighted}
\label{fusion}
\begin{tabular}{l | c  c c  c c c c c c c}
\toprule
Methods& Additive  &MP   &Gated  &MB &GMU &GMB &Concat &MRN &DMRN &DMRFE\\
\midrule
Early Fusion&59.9 &67.9 &67.9&69.2&\textbf{70.5}&\textbf{70.2}&61.0 &69.8 &68.0&-\\
\midrule
Late Fusion &71.3 &71.4 &70.5 &70.5 &71.6 & 71.0 &72.7&70.8&\textbf{73.1}&\textbf{73.3}\\
\midrule
Decision Fusion&\textbf{70.5}&64.5&65.2&64.6&67.6&67.3&69.7&63.8&\textbf{70.4} &-\\
\bottomrule
\end{tabular}
\end{center}
\vspace{-5mm}
\end{table}

\vspace{2mm}
\noindent \textbf{Audio-Visual Fusion.} 
Table \ref{fusion} shows audio-visual event localization prediction accuracy of different multimodal feature fusion methods on AVE dataset. Our DMRN model in the late fusion setting can achieve better performance than all compared methods, and our DMRFE model can further improve performance. 
We also observe that late fusion is better than early fusion and decision fusion. The superiority of late fusion over early fusion demonstrates that temporal modeling before audio-visual fusion is useful. We know that auditory and visual modalities are not completely aligned, and the temporal modeling can implicitly learn certain alignments between the two modalities, which is helpful for the audio-visual feature fusion task. The decision fusion can be regard as a type of late fusion but using lower dimension (same as the category number) features. The late fusion outperforms the decision fusion, which validates that processing multiple features separately and then learning joint representation using a middle layer rather than the bottom layer is an efficient fusion way.   

\vspace{1mm}
\noindent \textbf{Full and Weak Supervision.}
Obviously, supervised event localization models are better than weakly supervised ones, but quantitative comparisons show that weakly-supervised approaches achieve promising event localization performance, which demonstrates the effectiveness of the MIL networks on address this task, and validates that the audio-visual event localization task can be addressed even in a noisy condition.

\begin{wraptable}{r}{0.42\textwidth}
\vspace{-12mm}
\begin{center}
\caption{Accuracy on cross-modality localization. A2V: visual localization from audio segment query; V2A: audio localization from visual segment query}\label{cml}
\begin{tabular}{c | c c}
\toprule  
Models  & $\mathrm{AVDLN}$ &$\mathrm{DCCA}$  \\\midrule
A2V &\textbf{44.8} & 34.8\\  \midrule
V2A &\textbf{35.6} & 34.1\\  \bottomrule
\end{tabular}
\end{center}
\end{wraptable} 
% \begin{table}
% \setlength\belowcaptionskip{5pt}
% \begin{center}
% \caption{Prediction accuracy on cross-modality localization. A2V: video localization from an audio segment input; V2A: audio localization from a video segment input}
% \label{cml}
% \begin{tabular}{c | c c | c | c c}
% \toprule
% Models  & $\mathrm{AVDLN}$ &$\mathrm{DCCA}$ &Models  & $\mathrm{AVDLN}$ &$\mathrm{DCCA}$\\ 
% \midrule
% A2V &  \textbf{44.78} &34.82 &V2A & \textbf{35.57} & 34.08\\
% \bottomrule
% \end{tabular}
% \end{center}
% \end{table}
% \vspace*{-\baselineskip}
% \vspace*{-\baselineskip}
%------------------------------------------
%\subsection{Cross-Modality localization Results}

\vspace{1mm}
\noindent \textbf{\noindent Cross-Modality Localization.}
Table \ref{cml} reports the prediction accuracy of our method and DCCA \cite{andrew2013deep} on cross-modality localization task.  Our AVDL outperforms DCCA over a large margin both on A2V and V2A tasks. Even using the strict evaluation metric (which counts only the exact matches), our models on both subtasks: A2V and V2A, show promising results, which further demonstrates that there are strong correlations between audio and visual modalities, and it is possible to address the cross-modality localization for unconstrained videos. 

% \subsection{Discussion (\textbf{Maybe we can delete this part})}
% Arandjelovic and Zisserman \cite{arandjelovic2017objects} formulated a MIL framework and designed a audio-visual correspondence learning network on a AudioSet-Instruments
% dataset to explore audio-visual object localization with showing impressive results. They shown that the correspondence scores learned from the network can successfully indicate location of sounding objects, and our audio-guided visual attention can also capture spatial positions of sound sources in unconstrained videos. Therefore, there is a potential to address spatial-temporal localization problem in a unified framework. Moreover, in this work, we consider segment-level temporal localization with spatial attention. It is very interesting to further study frame-level localization.     

%------------------------------------------
\section{Conclusion}
\label{sec:conclusion}
In this work, we study a suit of five research questions in the context of three audio-visual event localization tasks. We propose both baselines and novel algorithms to address each of the three tasks. Our systematic study well supports our findings: modeling jointly over auditory and visual modalities outperforms independent modeling, audio-visual event localization in a noisy condition is still tractable, the audio-guided visual attention is able to capture semantic regions of sound sources and can even distinguish audio-visual unrelated videos, temporal alignments are important for audio-visual feature fusion, the proposed dual residual network is capable of audio-visual fusion, and strong correlations existing between the two modalities enable cross-modality localization.  
\section{Acknowledgement}
This work was supported by NSF BIGDATA 1741472. We gratefully acknowledge the gift donations of Markable, Inc., Tencent and the support of NVIDIA Corporation with the donation of the GPUs used for this research. This article solely reflects the opinions and conclusions of its authors and neither NSF, Markable, Tencent nor NVIDIA.
%\clearpage

\bibliographystyle{splncs}
\bibliography{egbib}

%% file: supp.tex
\pagestyle{headings}
\mainmatter
\title{Supplementary File \\ Audio-Visual Event Localization in Unconstrained Videos } % Replace with your title
\titlerunning{Supplementary File}
\authorrunning{Y. Tian, J. Shi, B. Li, Z. Duan, and C. Xu}
\author{
       }
\institute{
}

\maketitle

\noindent In this material, firstly, we show how we gather the \emph{Audio-Visual Event} (AVE) dataset in Sec. \ref{ave}. Then we describe the implementation details of our algorithms in Sec. \ref{detail}. Finally, we provide additional experiments in Sec. \ref{ex}.

%\noindent Additionally, we show a demo video of the AVE dataset and our results in Demo$\_$ID$\_$1818.mp4.

\section{AVE: The \emph{Audio-Visual Event} Dataset}
\label{ave}

Our \emph{Audio-Visual Event} (AVE) dataset contains 4143 videos covering 28 event categories. The video data is a subset of AudioSet \cite{gemmeke2017audio} with the given event categories, based on which the temporal boundaries of the audio-visual events are manually annotated.
%-----------------------------------------
\subsection{Gathering and Preparing Dataset}

With the proliferation of video content,
YouTube becomes a good resource for finding unconstrained videos.
The AudioSet \cite{gemmeke2017audio} released by Google is a large-scale audio-visual dataset that contains 2M 10-second video clips from Youtube. Each video clip corresponds to one of the total 632 event labels that is manually-annotated to describe the audio event.
In general, the events cover a variety of category types such as human and animal sounds, musical instruments and genres, and common everyday environmental sounds.
Although the videos in AudioSet contain both audio and visual tracks, a lot of them are not suitable for the audio-visual event localization task. For example, visual and audio content can be completely unrelated (\emph{e.g.}, train horn but no train appears, wind sound but no corresponding visual signals, the absence of audible sound, \emph{etc}).

To prepare our dataset, we select 34 categories including around $10,000$ videos from the AudioSet. Then we hire trained in-house annotators to select a subset of them as the desired videos, and further mark the start and end time at a resolution of 1 second as the temporal boundaries of each audio-visual event. We set a criterion that all annotators followed in the annotation process: a desired video should contain the given event category for at least a two-seconds-long segment from the whole video, in which the sound source is visible \emph{and} the sound is audible. This results in total 4143 desired videos covering a wide range of audio-visual events (\emph{e.g.}, woman speaking, dog barking, playing guitar, and frying food, \emph{etc.}) from different domains \emph{e.g.}, human activities, animal activities, music performances, and vehicle sounds.

\section{Implementation Details}
\label{detail}
Videos in AVE dataset are divided into training (3339), validation (402), and testing (402) sets. For supervised and weakly-supervised audio-visual event localization tasks, we randomly sample videos from each event category to build the train/val/test datasets. For evaluating the cross-modality localization performance, we only sample testing videos from short-event videos (events in these videos are all strictly smaller than the total 10s duration). We implement our models using Pytorch \cite{pytorch} and Keras \cite{chollet2015keras} with Tensorflow \cite{tensorflow2015-whitepaper} as backend. Networks are optimized by Adam \cite{kingma2014adam}. The LSTM hidden state size and contrastive loss margin are set to $128$ and $2.0$, respectively.

\section{Additional Experiments}
\label{ex}
Here, we compare different supervised audio-visual event localization models with different features in Sec. \ref{st}. The audio-visual event localization results with different attention mechanisms are shown in Sec. \ref{dam}.
%%%%
\setlength{\tabcolsep}{1pt}
\begin{table}
\caption{Supervised  audio-visual event localization prediction accuracy ($\%$) of each event category on AVE test dataset. A, $\mathrm{V_s}$, $\mathrm{V_{c3d}}$, $\mathrm{V_{s+c3d}}$, A$\mathrm{V_s}$, A$\mathrm{V_{c3d}}$, and A$\mathrm{V_{s+c3d}}$ refer to supervised audio, spatial, C3D, spatial + C3D, audio + spatial, audio + C3D, audio + spatial + C3D features-based models, respectively. Notice that the $\mathrm{V_s}$ model denotes the V model in our main paper. With additional C3D features, the $\mathrm{AV_{s+c3d}}$ model does not show noticeable improvements than  the $\mathrm{AV_s}$ model over all event categories. So, we only utilize spatial visual features in our main paper. The top-2 results are highlighted in bold}
\begin{center}
\resizebox{\textwidth}{!}{
\begin{tabular}{l | c c c c c c c c c c c c c c}

\toprule
 Models& bell  & man   &dog  &plane &car &woman &copt. &violin &flute & ukul. &frying  &truck  &shofar &moto. \\

\midrule
A &83.9&54.1&49.4&51.1&40.0&36.5&44.1&66.1&81.8&\textbf{78.1}&77.8&20.0&\textbf{61.0}& 34.4 \\
\hline
$\mathrm{V_s}$ & 76.7&40.6&44.1&68.3&\textbf{60.6}&24.7&50.6&44.4&44.7&17.5&70.6&69.2&40.0&66.7  \\
$\mathrm{V_{c3d}}$ &61.7 &33.5 &38.2 &\textbf{77.2} &57.2 &36.4 &55.3&40.0&23.5&14.4 &53.3 &42.3 &48.0 &70.0\\
$\mathrm{V_{s+c3d}}$ &76.7 &41.2&38.8&\textbf{77.2}&60.0&51.2&57.1&58.3&40.0&42.5&75.6&\textbf{80.0}&60.0 &72.2\\
\hline
$\mathrm{AV_s}$ &\textbf{84.4}&\textbf{57.6}&\textbf{55.3}&\textbf{77.2}&56.7&\textbf{72.4}&53.5&\textbf{80.6}&87.6&\textbf{80.0}&\textbf{80.0}&75.4&60.0&68.9\\
$\mathrm{AV_{c3d}}$ &83.3&\textbf{62.9}&53.5&72.8&49.4&\textbf{81.8}&\textbf{61.2}&\textbf{72.2}&\textbf{88.2}&73.8&\textbf{80.0}&40.0&\textbf{62.0}&\textbf{74.4}\\
$\mathrm{AV_{s+c3d}}$&\textbf{85.0}&50.6&\textbf{57.1}&\textbf{76.1}&\textbf{66.7}&71.2&\textbf{67.1}&71.2&\textbf{90.6}&75.6&\textbf{85.6}&\textbf{78.5}&\textbf{62.0}&\textbf{73.3}\\
\toprule
 Models&  guitar& train& clock   &banjo  &goat &baby &bus &chain. &cat &horse &toilet  &rodent  &acco. &mand.\\
 \midrule
A&\textbf{70.6}&65.3&81.3&\textbf{84.4}&53.0&61.3&8.3&68.1&30.0&8.3&70.6&49.0&60.7&64.7 \\
\hline
$\mathrm{V_s}$&57.8&73.5 &79.4&45.6&62.0&51.3 &\textbf{60.0}&73.1&23.3&\textbf{35.0}&60.6&42.0&66.0&41.3\\
$\mathrm{V_{c3d}}$&57.8 &77.1&78.1&40.6&57.0&17.5&43.3&43.1&11.7&13.3&72.8&9.0&34.0&22.7\\
$\mathrm{V_{s+c3d}}$ &48.9&68.8&66.3&61.7&\textbf{72.0}&20.0&\textbf{56.7}&73.8&21.7&20.0&71.1&48.0&64.0&39.3\\
\hline
$\mathrm{AV_s}$& 63.9&\textbf{88.8}&81.3&76.1&\textbf{75.0}&57.5&41.7&\textbf{83.1}&\textbf{61.7}&\textbf{33.3}&\textbf{83.9}&\textbf{57.0}&\textbf{74.7}&63.3\\
$\mathrm{AV_{c3d}}$ &69.4&82.4&\textbf{88.8}&\textbf{79.4}&44.0&\textbf{68.8}&40.0&76.9&\textbf{38.3}&20.0&76.1&53.0&64.7&\textbf{72.7}\\
$\mathrm{AV_{s+c3d}}$ &\textbf{70.0}&\textbf{85.3}&\textbf{88.1}&67.8&60.0&\textbf{67.5}&5.0&\textbf{82.5}&33.3&18.3&\textbf{88.3}&\textbf{70.0}&\textbf{81.3}&\textbf{66.7}\\
\bottomrule
\end{tabular}
}
\end{center}
\label{SEL}
\vspace{-5mm}
\end{table}
%===============================================================================================
%\setlength{\tabcolsep}{1.2pt}
\begin{table}
\caption{Overall accuracy (\%) of supervised audio-visual event localization with different features on AVE test dataset}
\begin{center}
\begin{tabular}{c| c| c c c | c c c}
\toprule
Models  & A &$\mathrm{V_s}$ &$\mathrm{V_{c3d}}$&$\mathrm{V_{s+c3d}}$ &A$\mathrm{V_s}$  &$\mathrm{AV_{c3d}}$& $\mathrm{AV_{s+c3d}}$\\
\midrule
Accuracy &  59.5& 55.3 &46.4 &57.9 & \textbf{{71.4}} & 68.7 & \textbf{{71.6}}\\
\bottomrule
\end{tabular}
\end{center}
\label{AVERAGEDAC}
\end{table}
%===============================================================================================
\setlength{\tabcolsep}{2pt}
\begin{table}
\begin{center}
\caption{Audio-visual event localization overall accuracy ($\%$) on AVE dataset. A$'$, A$'$-att, V, V-att, A$'$+V, A$'$+V-co-att denote that these models use audio, attended audio, visual, attended visual, audio-visual, and attended audio and attended visual features, respectively. Note that V represents that the model only use spatial visual features extracted from VGGNet, and the models without attention use global average to produce feature vectors;  A$'$ models use audio features extracted from the last pooling layer of pre-trained VGG-like model in \cite{hershey2017cnn} (for details, please see Sec. \ref{dam}) }
\label{att}
\begin{tabular}{l | c c| c c |cc }
\toprule
 Models& A$'$  &A$'$-att &V    &V-att  &A$'$+V &A$'$+V-co-att\\
\midrule
Accuracy &54.3&54.1 &55.3&58.5&\textbf{70.2}&\textbf{69.9}\\
\bottomrule
\end{tabular}
\end{center}
\vspace{-5mm}
\end{table}

\begin{figure}[t!]
\begin{center}
  \includegraphics[width=\linewidth]{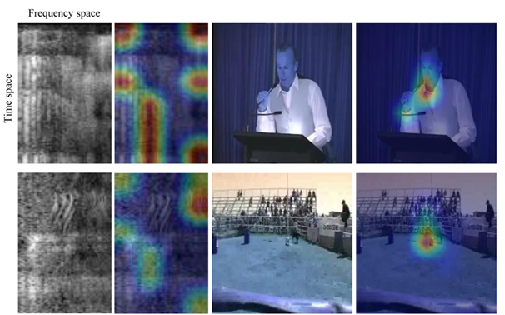}
\end{center}
   \caption{Visual results of visual-guided audio attention and audio-guided visual attention mechanisms. Each row represents one example. From left to right, images are log-mel spectrum patch, visual-guided audio attention map, a reference video frame, and audio-guided visual attention map, respectively.
   }
\label{fig:att}
\end{figure}
\subsection{Spatio-Temporal Feature for Audio-Visual Event Localization}
\label{st}
Although 2D CNNs pre-trained on ImageNet are effective in extracting high-level visual representations for static images, they fail to capture dynamic features modeling motion information in videos. To analyze whether temporal information is useful for the audio-visual event localization task, we utilize deep 3D convolutional neural network (C3D) \cite{tran2015learning} to extract spatio-temporal visual features. In our experiments, we extract C3D feature maps from $pool5$ layer of C3D network pre-trained on Sport1M \cite{karpathy2014large}, and obtain feature vectors by global average pooling operation. Tables \ref{SEL} and \ref{AVERAGEDAC} show supervised audio-visual event localization results of different features on AVE dataset.

Table \ref{AVERAGEDAC} shows the the overall accuracy on the AVE dataset. we see that A outperforms $\mathrm{V_{s}}$, both of them are better than $\mathrm{V_{c3d}}$ by large margins, and A$\mathrm{V_{s+c3d}}$ is only slightly better than A$\mathrm{V_{s}}$. It demonstrates that audio and spatial visual features are more useful to address the audio-visual event localization task than C3D features on the AVE dataset. From Table \ref{SEL}, we can find that $\mathrm{V_{c3d}}$ related models can obtain good results, only when videos have rich action and motion information (\emph{e.g.} plane, motocycle, and train \emph{etc}).

\subsection{Different Attention Mechanisms}
\label{dam}
In our paper, we propose an audio-guided visual attention mechanism to adaptively learn which visual regions in each segment of a video to look for the corresponding sounding object or activity. Here, we further explore visual-guided audio attention mechanism and audio-visual co-attention mechanism, where the latter integrates audio-guided visual attention and visual-guided audio attention. These attention mechanisms serve as a weighted global pooling method to generate audio or visual feature vectors. The visual-guided audio attention function is similar to that in the audio-guided visual attention model, and the co-attention model uses both attended audio and attended visual feature vectors.

To implement visual-guided audio attention mechanism, we extract audio features from the last pooling layer of pre-trained VGG-like model in \cite{hershey2017cnn}. Note that the network uses a log-mel spectrogram patch with 96 $\times$ 64 bins to represent a 1s waveform signal, so its pool5 layer will produce feature maps with spatial resolution;
this is different than audio features of A models in our main paper and in Tabs.~\ref{SEL} and~\ref{AVERAGEDAC} of this supplementary file. The reason is that the audio features in A models are 128-D vectors extracted from the last fully-connected layer. We denote a model using audio features in this section as A$'$ to differentiate it from the model A used in our main paper and in Tabs.~\ref{SEL} and~\ref{AVERAGEDAC}.

Table \ref{att} illustrates supervised audio-visual event localization results of different attention models. We can see that the the A$'$ model in Tab. \ref{att} is worse than the A model in Tab. \ref{AVERAGEDAC}, which demonstrates that the audio features extracted from the last FC layer of \cite{hershey2017cnn} is more powerful. Similar to results in our main paper, V-att outperforms V. However, A$'$-att is not better than A$'$, and A$'$+V-co-att is slightly worse than A$'$+V, which validate that visual-guided audio attention and audio-visual co-attention can not effectively improve audio-visual event localization performance.
Figure \ref{att} illustrates visual results of audio attention and visual attention mechanisms. Clearly, we can find that audio-guided visual attention can locate semantic regions with sounding objects. We also observe that the visual-guided audio attention tends to capture certain frequency patterns, but it is pretty hard to interpret the results of visual-guided audio attention, which we leave to explore in the future work.

%\clearpage

\bibliographystyle{splncs}
\bibliography{supp_bib}